\documentclass[10pt,twocolumn,letterpaper]{article}

\usepackage{times}
\usepackage{epsfig}
\usepackage{graphicx}
\usepackage{amsmath}
\usepackage{amssymb}

\usepackage{subfig,multirow,multicol,array}
\usepackage{amsfonts,amssymb}
\usepackage{algorithmic}
\usepackage[ruled]{algorithm2e}	

\usepackage{verbatim}
\usepackage[usenames,dvipsnames]{color}
\usepackage{soul}

\usepackage[pagebackref=true,breaklinks=true,letterpaper=true,colorlinks,bookmarks=false]{hyperref}

\def\cvprPaperID{5324} % *** Enter the CVPR Paper ID here

\usepackage{colortbl}
\definecolor{mygray}{gray}{.9}

\begin{document}

%%%%%%%%% TITLE
\title{Stratified Labeling for Surface Consistent Parallax Correction \\and Occlusion Completion}

\author{Jie Chen$^1$, Lap-Pui Chau$^1$, and Junhui Hou$^2$\\
\small $^1$ST Engineering - NTU Corporate Lab, Nanyang Technological University, Singapore\\
\small $^2$Department of Computer Science, City University of Hong Kong\\
%\and
%Department of Computer Science, City University of Hong Kong\\
}

\maketitle
%\thispagestyle{empty}

%%%%%%%%% ABSTRACT
\begin{abstract}
	The light field faithfully records the spatial and angular configurations of the scene, which facilitates a wide range of imaging possibilities. 
	In this work, we propose an LF synthesis algorithm which renders high quality novel LF views far outside the range of angular baselines of the given references. 
	A stratified synthesis strategy is adopted which parses the scene content based on stratified disparity layers and across a varying range of spatial granularities. Such a stratified methodology proves to help preserve scene structures over large perspective shifts, and it provides informative clues for inferring the textures of occluded regions. A Generative-Adversarial network model is further adopted for parallax correction and occlusion completion conditioned on the stratified synthesis features. Experiments show that our proposed model can provide more reliable novel view synthesis quality at large baseline extension ratios. Over 3dB quality improvement has been achieved against state-of-the-art LF view synthesis algorithms.
\end{abstract}\vspace{-0.2cm}

\section{Introduction}\label{sec_introduction}

With the commercialization of light field cameras such as Lytro \cite{Ng2005} and Raytrix \cite{perwass2012single}, light field imaging has become a popular topic and is attracting extensive research and industrial attentions. The light field (LF) is a vector function that describes the amount of light propagating in every direction through every point in free space \cite{lippmann1908la}. Compared with conventional 2D cameras, LF cameras can capture extra directional information for each light ray, and such information enables exciting applications such as refocusing, 3D scene reconstruction \cite {Kim2013, perra2016analysis}, material recognition \cite{wang20164d}, reflection/specularity removal \cite{ni2018reflection, tao2016depth}, and virtual/augment reality display \cite{Huang2015LFStereo}, to just name a few. 

One of the promising applications for LF imaging is in real-footage acquisition and immersive display of 3D environments. Compared with conventional stereo system which uses the binocular disparity for the brain to perceive the depth, LF allows to display correct scene perspectives from any viewing angles with realistic presentation of details such as shading, specularity, and focus shift. With the latest development of LF display technology \cite{stern2014perceivable,huang2015light} that aims at achieving glasses-free and fatigue-free immersive 3-D perception, LF is now considered as a promising future media for 3-D telepresence and virtual/augmented/mixed reality applications. 
One of the greatest technical challenges for these applications, is the extremely bulky size of LF data, which requires a great effort in the data acquisition, and a large bandwidth for transmission. 
This gives rise to a lot of research efforts focused on compact representation and compression technologies of the LF data \cite{hou2018light, chen2018light}. 

Another perspective of attacking this limitation is to synthesize novel LF views via a post-processing algorithm given small amount of LF information. Research works are seen on the interpolation of densely sampled LF views based on a sparse set of inputs \cite{kalantari2016learning,wang2017light,wu2017light}. In-between views could be smoothly synthesized which greatly reduces the amount of data required for transmission at real-time scenarios. 

Besides view interpolation, the concept of view extrapolation or baseline extension presents another promising direction. The need for an expanded LF/camera baseline is multi-fold. First, it gives a wider viewing angle which provides more visual information to the viewer. A larger baseline facilitates subsequent computational procedures: higher sub-pixel precision could be achieved for disparity estimation; more dramatic refocusing effect could be achieved. Most importantly, it gives more free-viewing angles for immersive display applications, without increasing data transmission bandwidth.

%%%%%%%%%%%%%%%%%%%%%%%Figure%%%%%%%%%%%%%%%%%%%%%%%
\begin{figure*}
	\begin{center}
		\includegraphics[width=0.95\linewidth]{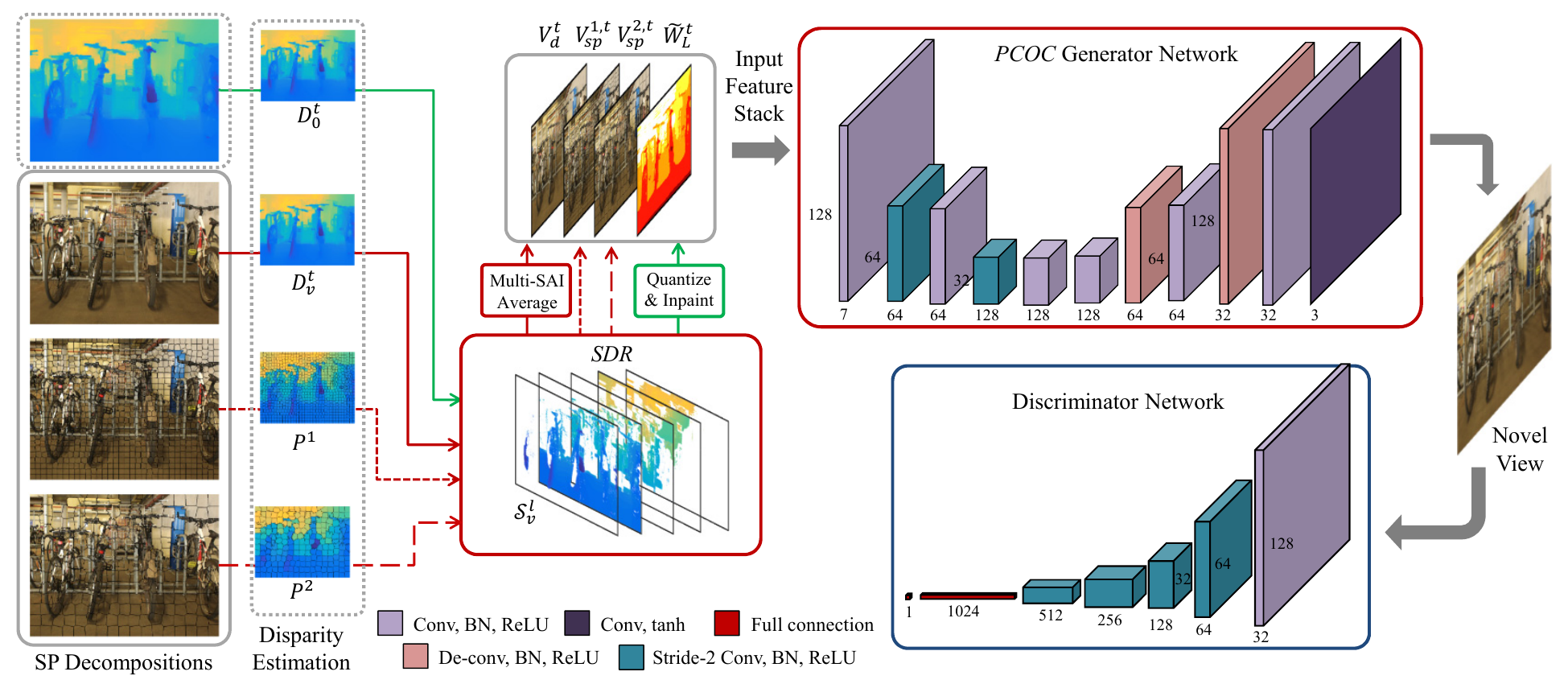}
	\end{center}
	\vspace{-0.5cm}
	\caption{System diagram for the proposed \textit{PCOC} framework.}
	\label{fig_system}
\end{figure*}
%%%%%%%%%%%%%%%%%%%%%%%Figure%%%%%%%%%%%%%%%%%%%%%%%

In this work, we will work on the problem of synthesizing novel LF views that is \textit{far} outside of the current LF baseline. We propose a stratified synthesis strategy which parses the scene content based on stratified disparity layers and across a varying range of spatial granularities. Such stratified methodology proves to be helpful in preserving scene content structures over large perspective shifts, and it provides informative clues for inferring the textures of occluded regions. % A Generative-Adversarial network \cite{goodfellow2014generative} model has been adopted for parallax correction and occlusion completion. Experiments show that our proposed model can provide more reliable novel view synthesis quality especially at large baseline extension ratios. Over 3dB quality improvement has been achieved against state-of-the-art LF view synthesis algorithms.

\section{Related Work}\label{sec_related}

\textbf{Light Field View Interpolation.}
Closely related to the problem of LF baseline extension, LF view interpolation aims at synthesizing the in-between LF views based on a sparse set of reference input views. 
Methods for LF view interpolation can be divided into two categories: 
The first category requires explicit estimation of the scene disparity to guide the view synthesis \cite{wang2015occlusion,chen2018accurate, kalantari2016learning}. Notably, Kalantari et al. \cite{kalantari2016learning} proposed to use the disparity estimated by a convolutional neural network (CNN) as guide to warp all reference views to the target angle, these warped views are subsequently fed into a second CNN for color refinement until a final prediction is reached. Bicubic interpolation was adopted during warping which links the gradients of errors from the synthesized views with those from the disparity estimation, which enables end-to-end training. Methods under this category is either limited by the precision of disparity estimation, or by the warping operators which only samples over a limited local area and fails to back-propagate synthesis errors over larger ranges.

Methods in the second category directly synthesize the target view via exploration of the geometrical features embedded within the EPIs \cite{wu2017light}, across stacked sub-aperture views along different directions \cite{shin2018epinet}, or via alternating filtering between the spatial-angular domains \cite{yeung2018fast}. Although methods under this category generally produce more realistic renderings, their performance is still limited for scenarios with large camera baselines.

% have achieved superior performance over the traditional approaches . Notably, Kalantari et al. \cite{kalantari2016learning} proposed a sequential convolutional neural network (CNN) with disparity estimation and Wu \textit{et al.} \cite{wu2017light} proposed to use a blur-deblur scheme to counter the problem of information asymmetry between angular and spatial domain and a single CNN is used to map the blurred Epipolar-Plane Images (EPIs) from low to high resolution.

\textbf{Parallax Magnification via Layered Depth Images}
Scene content at different depth show complex occlusion relationships when viewed from distant viewing angles. 
It is a well-adopted strategy to segment the pixels into separate layers based on their motion \cite{Liu2005motion} and disparity range \cite{he1998layered,Zhou2018stereo} for independent processing. Notably, Zhou et al. \cite{Zhou2018stereo} proposed an end-to-end training framework that first leans the Multi-Plane Images (MPI) with increasing depth ranges. The inferred MPIs are then used to synthesize a range of novel views via a subsequent module.
Such a framework preserves the scene geometry in an computationally efficient way, however it produces noticeable distortions when the parallax shift is large; and it is beyond its capability to deal with large ambiguous regions caused by occlusion.

% We present motion magnification, a technique that acts like a microscope for visual motion. It can amplify subtle motions in a video sequence, allowing for visualization of deformations that would otherwise be invisible. To achieve motion magnification, we need to accurately measure visual motions, and group the pixels to be modified. After an initial image registration step, we measure motion by a robust analysis of feature point trajectories, and segment pixels based on similarity of position, color, and motion. A novel measure of motion similarity groups even very small motions according to correlation over time, which often relates to physical cause. An outlier mask marks observations not explained by our layered motion model, and those pixels are simply reproduced on the output from the original registered observations.
% The motion of any selected layer may be magnified by a userspecified amount; texture synthesis fills-in unseen “holes” revealed by the amplified motions. The resulting motion-magnified images can reveal or emphasize small motions in the original sequence, as we demonstrate with deformations in load-bearing structures, subtle motions or balancing corrections of people, and “rigid” structures bending under hand pressure.

\textbf{Context-Aware Image Inpainting.} The challenge of predicting ambiguous image content caused by occlusion is directly related to the problem of image synthesis and inpainting. The Generative-Adversarial Network (GAN) \cite{goodfellow2014generative} show wonderful performance for these applications.
Semantic labels have been used to provide content consistent outcomes for image synthesis \cite{isola2017image,wang2017high}. Global and local context features have been adopted to provide a more stable and consistent inpainting \cite{iizuka2017globally,yeh2017semantic}. These works provide valuable inspirations to our work, in which we aim at inpainting the occluded regions with surface consistent constraints.

In this work, we aim at the task of synthesizing high quality novel LF views \textit{far} outside the range of angular baselines of the given reference LF.
Different from LF view interpolation or near field view extrapolation, this is a more challenging problem in the following sense:
\begin{enumerate}
	\item The variations between the target view and the given references are much larger. A typical LF view interpolation problem involves pixel translation/scaling in the order of several pixels; however baseline extension involves translation/scaling in the order of more than 10 pixels. Even small inaccuracy in the disparity estimation could lead to obvious distortions.
	\item The occlusion relationships among the scene content need to be more accurately and robustly modeled. Objects at different distances from the camera will have dramatically different occlusion relationships when the viewing angle is significantly changed.
	\item The handling of occluded areas is much more challenging as significant changes in viewing angles could expose large un-seen areas from the reference views. How to infer the content of these occluded area based on the spatial/angular context is an important issue to be addressed. 
\end{enumerate}

\section{Proposed Method}

The LF describes the intensity and direction of every light ray that propagates within an imaging system, and it is usually parameterized as a 4D variable $\mathcal{L}(x,y,s,t)$: with $(x,y)$ representing the spatial domain, while $(s,t)$ representing the angular domain \cite{Ng2005}. 
In this work, we limit the LF angular domain $(s,t)$ to be along horizontal direction for the purpose of simplifying computational complexity. However, extension to other angular directions or higher angular dimensionality is straightforward.

Suppose we have a densely sampled LF $L(x,y,s)$ with a limited angular baseline, which is spanned by an horizontal array of $(2\cdot M+1)$ Sub-Aperture Images (SAI) \cite{Ng2005}. These SAIs look at the scene from slightly different, equidistant viewing angles \cite{chen2018light}.
We denote these SAIs as $\{I_v(x,y)|v=(-M,-M+1,\cdot\cdot\cdot,0,\cdot\cdot\cdot,M-1,M)\}$, with $v$ indicating the SAI location with respect to the central SAI $I_0(x,y)$. 

Given such a limited angular baseline of $[-M,M]$, the purpose of our framework is to expand the LF baseline, and synthesize novel SAIs far outside of the current angular limit:
\begin{equation}\label{eq_pbsetup}
I^t= \mathcal{R}[I_{-M},\cdot\cdot\cdot,I_{0},\cdot\cdot\cdot,I_{M}, t],~|t| \gg M.
\end{equation}
Here we use $\mathcal{R}$ to represent the mapping between the input LF reference SAIs $\{I_{-M},\cdot\cdot\cdot, I_M\}$ and the target novel SAI $I^t$. To deal with the challenges discussed in Sec. \ref{sec_related}, we propose a novel SAI synthesis strategy that significantly extends camera baselines based on a Generative-Adversarial Network (GAN). The framework benefits from combining informative geometrical clues via transformations based on stratified disparity layers and spatial granularities. 

The system diagram of our proposed framework is shown in Fig. \ref{fig_dpInpaint}, named \textit{Stratified Labeling for Surface Consistent parallax correction and occlusion completion} (\textit{SLSC}). 
The \textit{stratification} is implemented at two levels: 
first, the scene is divided into different disparity layers. For a target SAI $I^t$, reference SAIs are warped to the desired angle in a layer-wise manner. These warped layers are then fused which efficiently preserves the occlusion relationships.
Second, we segment the reference SAIs into different spatial granularities in forms of Superpixels (SP). Pixels within the same SP will be transformed identically, which provides \textit{un-distorted} content reference within the granularity level.
These informative clues from stratified transform operations will be combined and fed into to a generative network, which corrects parallax distortions, and completes occluded regions. The details of the proposed \textit{SLSC} framework will be discussed in the following.

\subsection{Stratified Disparity Rendering for Structure Projection}\label{sec_dpSeg}

\subsubsection{Projection over Stratified Disparity Layers}

With recent advancement of LF disparity estimation algorithms \cite{wanner2014variational,jeon2015accurate, wang2016occlusion,chen2018accurate}, given a densely sampled LF, an accurate disparity map with sub-pixel precision can be efficiently calculated. 
We adopt the algorithm by Chen et al. \cite{chen2018accurate} as our disparity estimator, since it shows the advantage in preserving the contours of complex occluding structures, which are most vulnerable to distortions over large perspective transforms. Additionally, the method also works on a superpixel granularity, which fits well into our granulated spatial operations.

We define the operator $\mathcal{W}[I, D]$ which warps the image $I$ according to the disparity map $D$. Bicubic interpolation \cite{Keys1981cubic} will be used for sub-pixel locations. 
Suppose we have a disparity estimation $D_v(x,y)$ for each reference LF SAI $I_v,~v\in[-M,M]$ using the method from \cite{chen2018accurate}. Along with $D_v(x,y)$, we also get an estimation confidence map $C_d(x,y)$.  
The target view $I^t(x,y)$ can be calculated based on the reference SAI $I_v(x,y)$ according to:
\begin{align}\label{eq_warping}
I_v^t(x,y)= &I_v[(x,y)+(t-v)\cdot D_v(x,y)],\\
=&\mathcal{W}[I_v(x,y), (t-v)\cdot D_v(x,y)].	
\end{align}
The subscript $v$ in $I_v^t(x,y)$ indicates the warping outcome is based on the input reference view $I_v$.

%To deal with the challenge of parallax distortion and ambiguity caused by occlusion, 
In order to preserve the scene geometry and its occlusion relationships over a significant angular transform, we implement a layer-wise warping and fusion scheme based on stratified disparity maps.
The idea of layer-wise warping has been implemented in earlier works \cite{Shade1998layered} \cite{Zhou2018stereo}. We further extend this strategy into a multi-granularity framework, which efficiently provides complementary information over distortion-prone areas.

We divide the disparity variation range of $D_v(x,y)$ into $L$ equal intervals. Pixels with disparity values that fall within the same interval are grouped as one layer. We denote these $L$ layer groups as $\{\Omega_v^l|l=1,2,\cdot\cdot\cdot,L\}$, with the superscript $l$ denoting the layer index. 
Consequently, a layered volume of disparity maps $\{\mathcal{S}_v^l|l=1,2,\cdot\cdot\cdot,L\}$ can be created according to:
\begin{equation}\label{eq_sd}
\mathcal{S}_v^l(x,y)=\begin{cases}
D_v(x,y), &(x,y)\in\Omega_v^l\\
0, &\text{otherwise}.
\end{cases}\\
\end{equation}
A visualization of $\mathcal{S}_v^l(x,y)$ can be seen in Fig. \ref{fig_system}.

With similar disparity values, it can be assumed that pixels of the same layer $\mathcal{S}_v^l$ will not alter occlusion relationships after angular transform. Warping to the target SAI $I_v$ is carried out for each layer independently according to:
\begin{equation}\label{eq_snt}
I_v^{t,l}=\begin{cases}
\mathcal{W}[I_v(x,y), (t-v)\cdot \mathcal{S}_v^l], &(x,y)\in\Omega_v^l\\
0, &\text{otherwise}.
\end{cases}\\
\end{equation}
Consequently, a \textit{stratified volume} of target SAIs $\{I_v^{t,l}| l=1,2,\cdot\cdot\cdot,L\}$ can be calculated.

\subsubsection{Layer Volume Fusion}\label{sec_fusion}

Given the layer-wise warped volume $\{I_v^{t,l}| l=1,2,\cdot\cdot\cdot,L\}$, we combine them to synthesize a complete target SAI.

Starting from disparity layer ($l=L$) that corresponds to the \textit{furthest} scene distance range, pixels from $\{I_v^{t,L}\}$ will occupy the corresponding pixels in $\tilde{I}_v^t$. Subsequently, nearer layers will be processed accordingly. When there is a pixel occupation conflict, nearer layers will replace the pixels from further layers. We use the tensor $R(x,y,l)$ to represent such fusion logic. For each pixel location $(x',y')$, only one layer of $R(x',y',l)$ along $l=1,2,\cdot\cdot\cdot,L$ are set to 1, with all the rest set to zeros. The non-zero layer corresponds to the layer closest to camera $\{I_v^{t,l}| l=1,2,\cdot\cdot\cdot,L\}$ with non-zero pixel contents. The final image can be calculated as:
\begin{equation}\label{eq_fusion}
\tilde{I}_v^t= \sum_{l}[I_v^{t,l}(x,y)\cdot R(x,y,l)],~l=1,2,\cdot\cdot\cdot,L.
\end{equation}

Here we define an operator $\mathcal{F}_L[I,D,t']$ which incorporates the above mentioned layer-wise warping plus layer fusion for the synthesis of novel SAIs, and we name such an operator \textit{Stratified Disparity Rendering} (\textit{SDR}).
$\mathcal{F}_L[I,D,t']$ functions to synthesize a novel view based on a reference SAI $I$, and its disparity map $D$. $t'$ indicates the angular shift distance for the target view (in units of angular distance between neighboring reference SAIs). The subscript $L$ indicates number of stratified layers.
Now the \textit{SDR} process defined in Eqs. (\ref{eq_sd}), (\ref{eq_snt}), and (\ref{eq_fusion}) can be combined and re-written as:
\begin{equation}
\tilde{I}_v^t= \mathcal{F}_L[I_v,D_v, t-v].
\end{equation}

Since we have one prediction $\tilde{I}_v^t$ based on every reference SAI $v=-M,\cdot\cdot\cdot,0,\cdot\cdot\cdot,M$. These predictions are largely similar, however they contain complimentary perspective information especially around the occlusion borders. They are combined to provide a more informative prediction according to:
\begin{equation}\label{eq_avg}
V_d^t= \sum_v{\tilde{I}_v^t}/(2\cdot M+1).
\end{equation}

\begin{figure}[t]
	\begin{center}
		\includegraphics[width=0.98\linewidth]{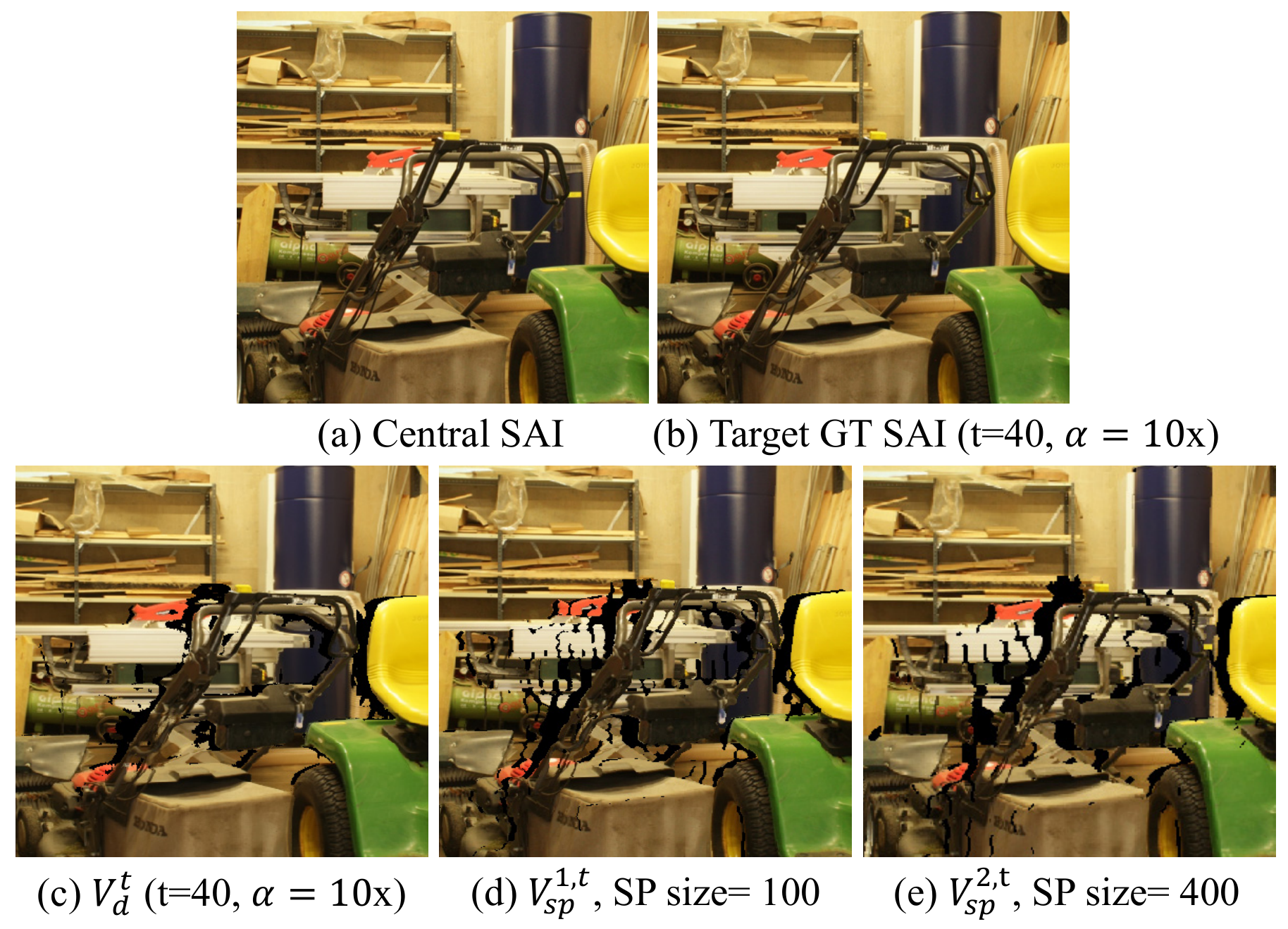}
	\end{center}
	\caption{(a) Central view for the LF data \textit{Workshop} \cite{Vamsi2017}, (b) shows the ground truth target SAI $t=40$, which is $\alpha=$10x times wider than the reference LF baseline. The warping outcome by the \textit{SDR} from different granularity levels $V_d^t$, $V_{sp}^{1,t}$, and $V_{sp}^{2,t}$ are shown in (c), (d), and (e), respectively.}
	\label{fig_warpComp}
\end{figure}

Fig. \ref{fig_warpComp}(c) shows an example of the rendered target view $V_d^t$ for the scene $Workshop$ from the MPI Light Field Archive \cite{Vamsi2017}, where $t=40$, with an angular baseline extension ratio of $\alpha$=10x. Compared with the ground truth target SAI in Fig. \ref{fig_warpComp}(b), obvious distortions and occlusion gaps can be observed.
% Due to unavoidable disparity estimation inaccuracies, there could show  especially when $t$ is large, as demonstrated in Fig. \ref{fig_warpComp}(a).

\begin{figure}[t]
	\begin{center}
		\includegraphics[width=1\linewidth]{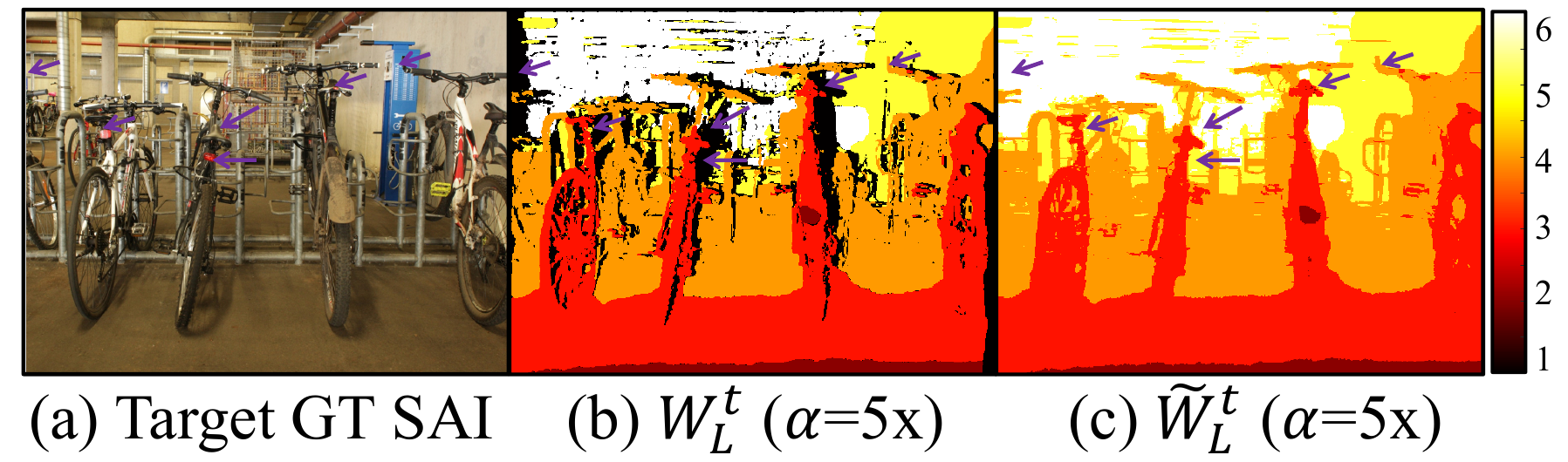}
	\end{center}
	\caption{(a) Ground truth target SAI $t$=20 ($\alpha$=5x) of scene \textit{Bikes} \cite{Vamsi2017}, (b) discretized disparity label map from \textit{SDR} for target SAI $t$=20, and (c) disparity label map $\tilde{W}_L^t$ inpainted from (b).}
	\label{fig_dpInpaint}
\end{figure}

Additionally, we can also calculate the disparity map for the target SAI with \textit{SDR}, in the same way as RGB images:
\begin{equation}
W^t(x,y)= \mathcal{F}_L[D_0(x,y),D_0(x,y),t].
\end{equation}
We further quantized $W^t(x,y)$ into $L$ discrete labels $W_L^t(x,y)$. An example of the labeled disparity map $W_L^t(x,y)$ for the scene \textit{Bikes} \cite{Vamsi2017} is shown in Fig. \ref{fig_dpInpaint}(b).

%%%%%%%%%%%%%%%%%%%%%%%Figure%%%%%%%%%%%%%%%%%%%%%%%
%%%%%%%%%%%%%%%%%%%%%%%Figure%%%%%%%%%%%%%%%%%%%%%%%

\subsection{Stratified Spatial Granularities for Distortion Correction}\label{sec_spSeg}

In order to prevent content distortion and preserve structural consistency over significant angular transforms, we propose to utilize \textit{SDR} across multiple spatial granularity levels based on units of superpixels (SP).

The concept of SP is to group pixels into perceptually meaningful atomic regions \cite{achanta2012slic,bergh2012seeds,li2015superpixel}. Boundaries of SP usually coincide with those of the scene content. The SPs are very adaptive in shape, and are more likely to segment uniform depth regions compared with rectangular units.
We implement SP segmentations on the central view $I_0$ at multiple SP scales (different SP scales correspond to different pixel numbers within each SP). Smaller SPs provide finer granularity that better binds to object boundaries, while larger granularity prevents distortion over complex, and textureless scene regions.

Based on the central view's pixel-wise disparity map $D_0(x,y)$, we calculate its SP-wise disparity map $P_0^s(x,y)$ as introduced in \cite{chen2018accurate}. The pixels within the same SP share identical disparity value (initially calculated as the median of confident pixels indicated by $C_d(x,y)$, and then regularized over inter-SP border pixel's discontinuities). The resulting SP-wise disparity map $P_0^s(x,y)$ (with superscript $s$ indicating the SP size) strictly adheres to the occluding boundaries, and preserves its internal pixel structure even after significant angular transform operations.

With the \textit{SDR} operator $\mathcal{F}$, we can synthesize the target novel SAI based on the SP-wise disparity map $P^s(x,y)$:
\begin{equation}
V_{sp}^{s,t}= \mathcal{F}_L{[I_0, P_0^s, t]}.
\end{equation} 
The central view can be warped to the target view based on different SP size levels, which provides predictions with pixel details strictly preserved at different scales. In our implementation, we set up 2 SP size levels with 100 and 400 pixels per SP, respectively. This gives synthesis outcome of $V_{sp}^{1,t}$, and $V_{sp}^{2,t}$. As shown in Figs. \ref{fig_warpComp}(d) and (e), compared with $V_d^{t}$, there are more occlusion gaps in $V_{sp}^{s,t}$. However, the SP helps correct regions with obvious wrong disparities over the occlusion borders (as explained in \cite{chen2018accurate}), and it keeps textureless surfaces geometrically consistent
after warping.

%%%%%%%%%%%%%%%%%%%%%%%Figure%%%%%%%%%%%%%%%%%%%%%%%
%%%%%%%%%%%%%%%%%%%%%%%Figure%%%%%%%%%%%%%%%%%%%%%%%

\subsection{Surface Consistent Features}

\begin{comment}
\begin{equation}
W_L^t(x,y)= l,~\text{if}~W^t(x,y)\in \Omega_k
\end{equation}
\end{comment}

As can be seen in Figs. \ref{fig_warpComp}(c), (d), (e) and \ref{fig_dpInpaint}(b), there exist large ambiguous areas (zero valued pixels) in the \textit{SDR} outputs $V_d^t$, $V_{sp}^{1,t}$, $V_{sp}^{2,t}$ and $W_L^t$. And we aim at in-painting these ambiguous areas with realistic predictions.

Since the discretized disparity label map $W_L^t$ has limited value ranges and much simpler distributions as compared with RGB images ($V_d^t$, $V_{sp}^{1,t}$, and $V_{sp}^{2,t}$), it is much easier to inpaint $W_L^t$. We adopt a classic image morphological dilation operator which propagates larger labels (which corresponds to more distant surfaces) within a local window to the zero marked ambiguous pixels. 
The dilation operator is based on the simple while efficient logic that \textit{any regions of ambiguity belongs to its furthest-away local surface}. An inpainted label map $\tilde{W}_L^t$ is shown in Fig. \ref{fig_dpInpaint}(c) for the data \textit{Bikes}. As can be seen, the missing areas are well completed with well-defined occluding contours.

With $\tilde{W}_L^t$ as \textit{surface consistent} guide, it can potentially instruct the inpainting process of RGB images by suggesting which surface the ambiguous region belongs to, such that relevant structural and textural features could be learned and transferred directly from these regions with identical labels and avoid confusion with the occluding surfaces.

%%%%%%%%%%%%%%%TABLE%%%%%%%%%%%%%%%%
\begin{table*}[ht]
	\small
	\centering\setlength\tabcolsep{1pt} % default value: 6pt
	\begin{center}		
		\caption{Quantitative evaluation and comparison for target SAIs synthesized by different algorithms at different baseline extension ratios.} 
		\label{tbl_psnr}
		\begin{tabular}{|>{\centering\arraybackslash}m{1.8cm}|>{\centering\arraybackslash}m{1.2cm}|>{\centering\arraybackslash}m{1.2cm}|>{\centering\arraybackslash}m{1.2cm}|>{\centering\arraybackslash}m{1.2cm}|>{\centering\arraybackslash}m{1.2cm}|>{\centering\arraybackslash}m{1.2cm}|>{\centering\arraybackslash}m{1.2cm}|>{\centering\arraybackslash}m{1.2cm}|>{\centering\arraybackslash}m{1.2cm}|>{\centering\arraybackslash}m{1.2cm}|>{\centering\arraybackslash}m{1.2cm}|>{\centering\arraybackslash}m{1.2cm}|>{\centering\arraybackslash}m{0.6cm}|>{\centering\arraybackslash}m{0.8cm}|>{\centering\arraybackslash}m{0.8cm}|>{\centering\arraybackslash}m{0.6cm}|>{\centering\arraybackslash}m{0.8cm}||}
			\hline 
			\multirow{2}{*}{\parbox{1cm}{\textit{Testing\\Data}}} &\multirow{2}{*}{\parbox{0.5cm}{\textit{Expan.\\Ratio}}}　&\multicolumn{2}{c|}{\textit{LVS-TOG16}\cite{kalantari2016learning}}&\multicolumn{2}{c|}{\textit{SAA-ECCV18} \cite{yeung2018fast}} &\multicolumn{2}{c|}{\textit{POBR-TIP18} \cite{chen2018accurate}} &\multicolumn{2}{c|}{\textit{SL (w/o SC)}}&\multicolumn{2}{c|}{\textit{SLSC}} \\
			\cline{3-12}
			&　&PSNR &SSIM &PSNR &SSIM &PSNR &SSIM &PSNR &SSIM  &PSNR &SSIM \\
			\hline\hline
			\multirow{3}{*}{\parbox{1cm}{\textit{Art\\Gallery2\\zoom}}} 
			&5x &19.02&0.82 &23.17&0.85 &22.60&0.89 &24.42&0.91 &28.35&0.92\\ 
			&7x &18.25&0.80 &19.76&0.81 &20.70&0.87 &22.92&0.89 &25.38&0.90\\
			&9x &17.13&0.76 &17.93&0.78 &19.36&0.85 &20.558&0.87 &23.28&0.87\\ \hline\hline
			\multirow{3}{*}{\parbox{1cm}{\textit{Bikes}}} 
			&5x &18.90&0.69 &23.75&0.59 &22.94&0.80 &25.37&0.86 &26.87&0.88\\ 
			&7x &18.49&0.64 &21.52&0.66 &21.23&0.74 &24.23&0.83 &25.48&0.83\\
			&9x &17.98&0.59 &19.93&0.73 &20.12&0.69 &22.54&0.78 &23.85&0.78\\ \hline\hline
			\multirow{3}{*}{\parbox{1cm}{\textit{Furniture1}}} 
			&5x &26.72&0.96 &34.56&0.96 &29.06&0.97 &30.12&0.97 &31.11&0.98\\ 
			&7x &26.45&0.95 &32.05&0.95 &28.12&0.96 &29.50&0.97 &30.29&0.97\\
			&9x &24.88&0.92 &28.91&0.93 &27.27&0.95 &28.44&0.96 &29.30&0.96\\ \hline\hline
			\multirow{3}{*}{\parbox{1cm}{\textit{Workshop}}} 
			&5x &19.89&0.78 &26.36&0.82 &24.32&0.91 &26.76&0.94 &30.30&0.95\\ 
			&7x &19.26&0.72 &23.29&0.75 &22.56&0.86 &25.82&0.93 &28.03&0.92\\
			&9x &18.00&0.65 &20.66&0.64 &21.32&0.82 &23.77&0.89 &26.46&0.89\\ \hline
			
		\end{tabular}\vspace{-0.5cm}
	\end{center}
\end{table*}
%%%%%%%%%%%%%%%TABLE%%%%%%%%%%%%%%%%

%%%%%%%%%%%%%%%%%%%%%%%Figure%%%%%%%%%%%%%%%%%%%%%%%
\begin{figure}[t]
	\begin{center}
		\includegraphics[width=0.8\linewidth]{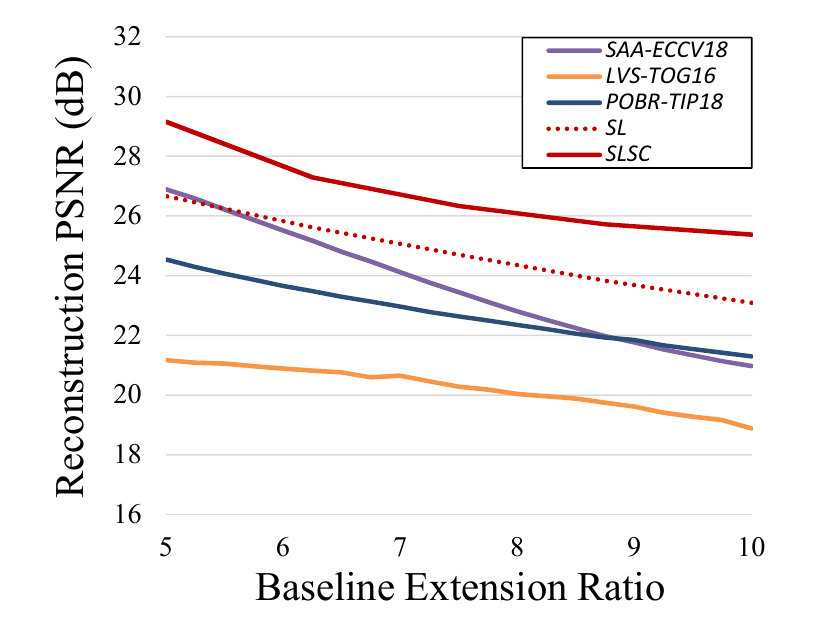}
	\end{center}
	\caption{Average PSNR of the synthesized target views at different baseline extension ratios ranging from $\alpha$=5x to 10x for the testing dataset.}
	\label{fig_psnr}
\end{figure}
%%%%%%%%%%%%%%%%%%%%%%%Figure%%%%%%%%%%%%%%%%%%%%%%%

\subsection{Generative-Adversarial Network for Parallax Correction and Occlusion Completion}
\subsubsection{The Parallax Correction and Occlusion Completion Network}

We propose a Generative-Adversarial Network to simultaneously correct parallax distortions caused by inaccurate disparity estimations, and inpaint the ambiguous regions caused by occlusion. As shown in Fig. \ref{fig_system}, the Parallax Correction and Occlusion Completion (\textit{PCOC}) network follows an encoder-decoder structure, which allows to reduce the memory usage and computational time by initially decreasing the resolution before further processing the image. The output is restored to the original resolution using deconvolution layers.

The initial warping outcomes by the \textit{SDR} from different granularity levels $\{V_d^t,V_{sp}^{1,t},V_{sp}^{2,t}\}$, along with the inpainted disparity label map $\tilde{W}^t_L$ are used as input to the \textit{PCOC} generator network in the following format:
\begin{align*}
\mathcal{T}_1(x,y)= &V_{sp}^{t,1}(x,y)- V_d^t(x,y),~\mathcal{T}_1\in\mathbb{R}^{W\times H\times 3},\\
\mathcal{T}_2(x,y)= &V_{sp}^{t,2}(x,y)- V_d^t(x,y),~\mathcal{T}_2\in\mathbb{R}^{W\times H\times 3},\\
\mathcal{T}_3(x,y)= &\tilde{W}_L^t(x,y)/L- 0.5,~\mathcal{T}_3\in\mathbb{R}^{W\times H\times 1}.
\end{align*}
Both $V_{sp}^{t,1}$ and $V_{sp}^{t,2}$ are subtracted with $V_d^{t}$, such that the input features $\mathcal{T}_1$ and $\mathcal{T}_2$ combine information from all granularity levels. With variation range around zero, $\mathcal{T}_1$ and $\mathcal{T}_2$ highlight the differences between the warping scales, which facilitate more efficient learning.  
$\mathcal{T}_3$ normalizes the label map $\tilde{W}^t_L$ to the range [-0.5,+0.5]. This feature is expected to guide the inpainting process of the RGB images to be \textit{surface consistent} with the occluded region.

The occlusion gaps are zero valued in $V_d^t$, $V_{sp}^{t,1}$, and $V_{sp}^{t,2}$. After the subtractions in $\mathcal{T}_1$ and $\mathcal{T}_2$, only those occlusion gaps that exist in both granularities are kept, which serves as an indicator to the \textit{PCOC} network concerning over which pixels to carry out \textit{occlusion completion}, and which areas to impose \textit{parallax correction} such that the final RGB output confirms to the statistical distribution of an undistorted natural image.

In our implementation, the spatial patch size is set as $W$=$H$=128. 
The dimensionality of the input feature to the \textit{PCOC} network is $\{\mathcal{T}_1,\mathcal{T}_2,\mathcal{T}_3\}\in\mathbb{R}^{W\times H\times 7}$, and the output is an RGB image $\mathring{I}^t\in\mathbb{R}^{W\times H\times 3}$. Note that in the end, $V_d^{t}$ will be added back to $\mathring{I}^t$ to restore color.

\subsubsection{Context Discriminator}

As illustrated in Fig. \ref{fig_system}, the context discriminator is based on a CNN which gradually compresses the image into small feature vectors. Outputs of the networks are continuous value corresponding to the probability of the \textit{PCOC} output being real. The discriminator works to improve the realism of the generator outputs by learning the distribution of ground truth views without parallax distortions or inpainting errors.

%%%%%%%%%%%%%%%%%%%%%%%Figure%%%%%%%%%%%%%%%%%%%%%%%
\begin{figure*}
	\begin{center}
		\includegraphics[width=0.98\linewidth]{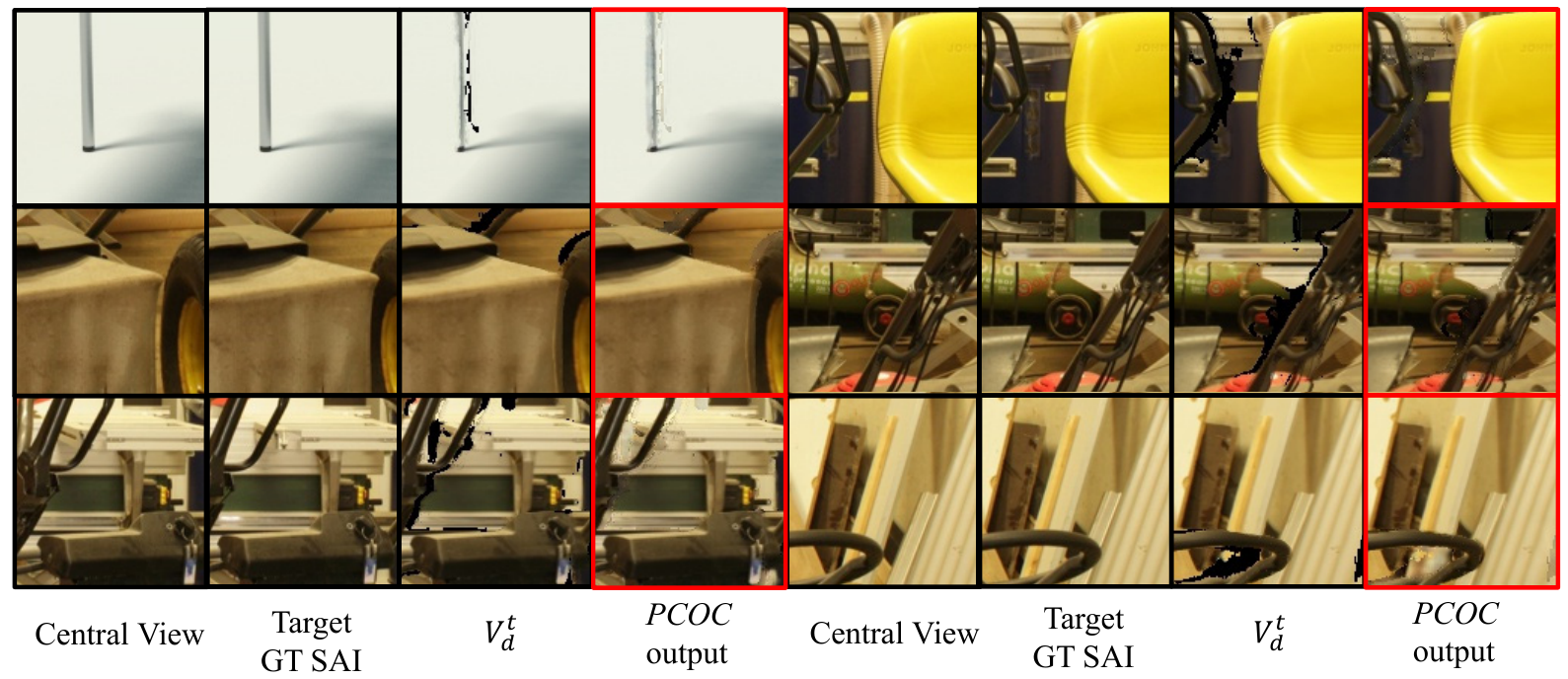}
	\end{center}\vspace{-0.4cm}
	\caption{Demonstration of \textit{PCOC}-Network outputs for selected testing data at baseline extension ratios between 5x and 9x.}
	\label{fig_pcocOutput}
\end{figure*}
%%%%%%%%%%%%%%%%%%%%%%%Figure%%%%%%%%%%%%%%%%%%%%%%%

\subsubsection{Training Details}

We used the MPI Light Field Archive \cite{Vamsi2017} for the training as well as evaluation of our model. 
The archive consists of 9 synthetic and 5 captured real-world scenes. All of the 14 LF data have a large angular base-line of 101 views, which allows enough angular flexibility to train our model. 9 center views (view index range from 47 to 55 out of the 101 views) were chosen to simulate the narrow-baseline LF input, side views with index range from 11 to 31 and 71 to 91 were used as ground truth for baseline extension. 2 real-world scenes \textit{Bikes} and \textit{Workshop}, and 2 synthetic scenes \textit{ArtGallery2zoom} and \textit{Furniture1} were used as the testing dataset. The rest of the 10 scenes in the MPI Light Field Archive were used as the training dataset, and they have been augmented according to the methods introduced in \cite{shin2018epinet}. Note that since the input LF baseline radius is $(9-1)/2=4$, therefore view 11 and view 71 corresponds to an angular baseline extension rate of $\alpha$=10x; and view 31 and 71 correspond to $\alpha$=5x, respectively.

In order to train the network such that inpainting of occlusion areas, and the parallax correction areas are realistic looking, two loss functions are jointly used: an $L_1$ loss term $\mathcal{L}_{L_1}$ for content fidelity between the rendered novel view $\mathring{I}^t$ and the ground truth view, and a Generative Adversarial Network (GAN) \cite{goodfellow2014generative} loss $\mathcal{L}_{GAN}$ to improve the realism of the results. The final objective is:
\begin{equation}
\mathcal{L}= \mathcal{L}_{L_1}+\lambda \mathcal{L}_{GAN}.
\end{equation}
To optimize our networks, We used stochastic gradient descent (SGD) to minimize the objective functions. We alternated between one gradient descent step on the generative \textit{PCOC} net, and then one step on the context discriminator net. Mini-batch size is set as 20 for better trade-off between speed and convergence. The Xavier approach \cite{glorot2010understanding} is used for network initialization, and the ADAM solver \cite{kingma2014adam} is adopted for system training, with parameter settings $\beta_1=$ 0.9, $\beta_2=$ 0.999, and learning rate $\alpha=$ 0.0001. Following suggestions from \cite{iizuka2017globally}, we train the \textit{PCOC} net 1000 epochs, and then start the alternating training with the discriminator.

\section{Model Evaluation}

We evaluate our \textit{SLSC} model both quantitatively and qualitatively, and compare its performance with state-of-the-art methods, including the spatial-angular alternating filtering algorithm (\textit{SAA-ECCV18}) \cite{yeung2018fast}, the learning based view synthesis algorithm (\textit{LVS-TOG16}) \cite{kalantari2016learning}, and a direct warping algorithm based on disparity maps estimated based on superpixel regularization over partially occluded border regions (\textit{POBR-TIP18}) \cite{chen2018accurate}. 
Note that both \textit{SAA-ECCV18} and \textit{LVS-TOG16} were originally designed for LF view interpolation given a sparsely sampled LF input. Both of them have been retrained over the MPI dataset for the purpose of baseline extension (extrapolation). 
4 interleaved layers have been adopted for \textit{SAA-ECCV18}. 9 adjacent center views (view index range from 47 to 55) were selected as the input reference for both networks, and side views of index range 11 to 31, and 71 to 91 were used as ground truth for training, which represents angular baseline extension limits from $\alpha=$5x to $\alpha=$10x. 

Another baseline method was also chosen for comparisons, in which we used $\{\mathcal{T}_1, \mathcal{T}_2\}$ as the input feature to the \textit{PCOC} network, without the surface consistent guide $\{\mathcal{T}_3\}$. This baseline was used to assess the importance of the surface consistency feature for the synthesis of novel views. We denote this method as \textit{SL} (without Surface Consistency feature \textit{SC}).

% 2 real-world scenes \textit{Bikes} and \textit{Workshop}, and 2 synthetic scenes \textit{ArtGallery2zoom} and \textit{Furniture1} were used as the testing dataset.

\subsection{Quantitative Evaluation}

We quantitatively evaluate the PSNR (Peak signal-to-Noise Ratio) and SSIM (Structural Similarity Index) of rendered novel views. Table \ref{tbl_psnr} shows the results for the four testing scenes at different baseline extension ratios of 5, 7, and 9 respectively.
Additionally, the average PSNR for all the testing data have been plotted as curves for each competing method in Fig. \ref{fig_psnr}. 

As can be seen, our proposed \textit{SLSC} model provides a consistent 2-5 dB advantage over all compared methods. The advantage is especially large at larger baseline extension ratios.

%%%%%%%%%%%%%%%%%%%%%%%Figure%%%%%%%%%%%%%%%%%%%%%%%
\begin{figure*}
	\begin{center}
		\includegraphics[width=0.8\linewidth]{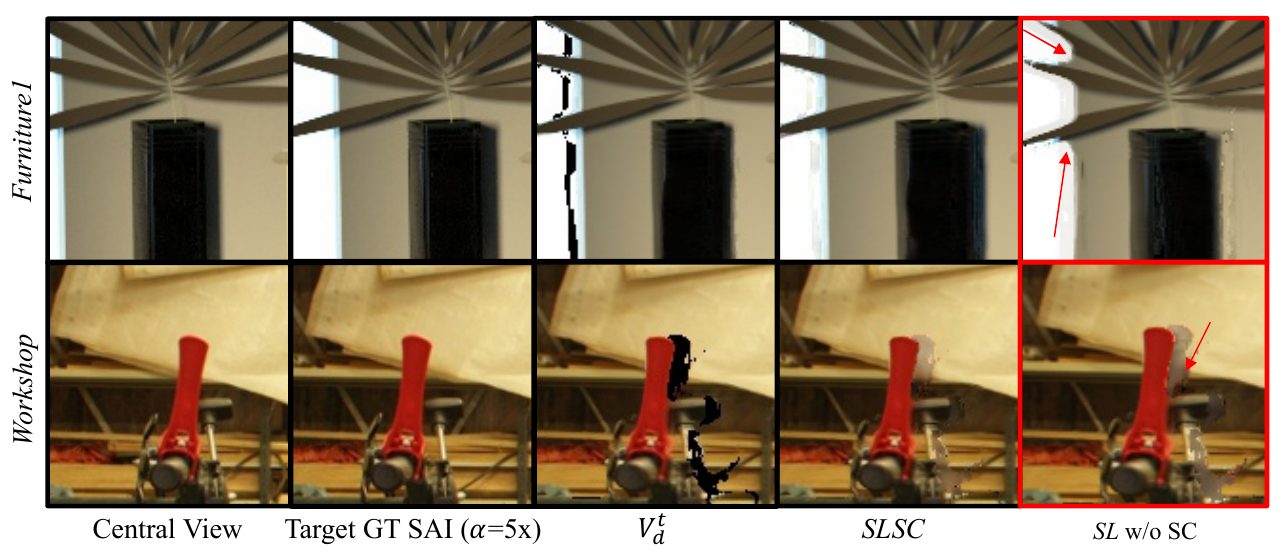}
	\end{center}
	\vspace{-0.6cm}
	\caption{Comparison of visual quality for synthesized target SAIs for the testing data \textit{Furniture1} ($t$=-20, $\alpha$=5x), and \text{Workshop} ($t$=20, $\alpha$=5x) from the proposed \textit{SLSC} model and the baseline model \textit{SL} (without surface consistent features).}
	\label{fig_scInfluence}
\end{figure*}
%%%%%%%%%%%%%%%%%%%%%%%Figure%%%%%%%%%%%%%%%%%%%%%%%

%%%%%%%%%%%%%%%%%%%%%%%Figure%%%%%%%%%%%%%%%%%%%%%%%
\begin{figure*}
	\begin{center}
		\includegraphics[width=0.98\linewidth]{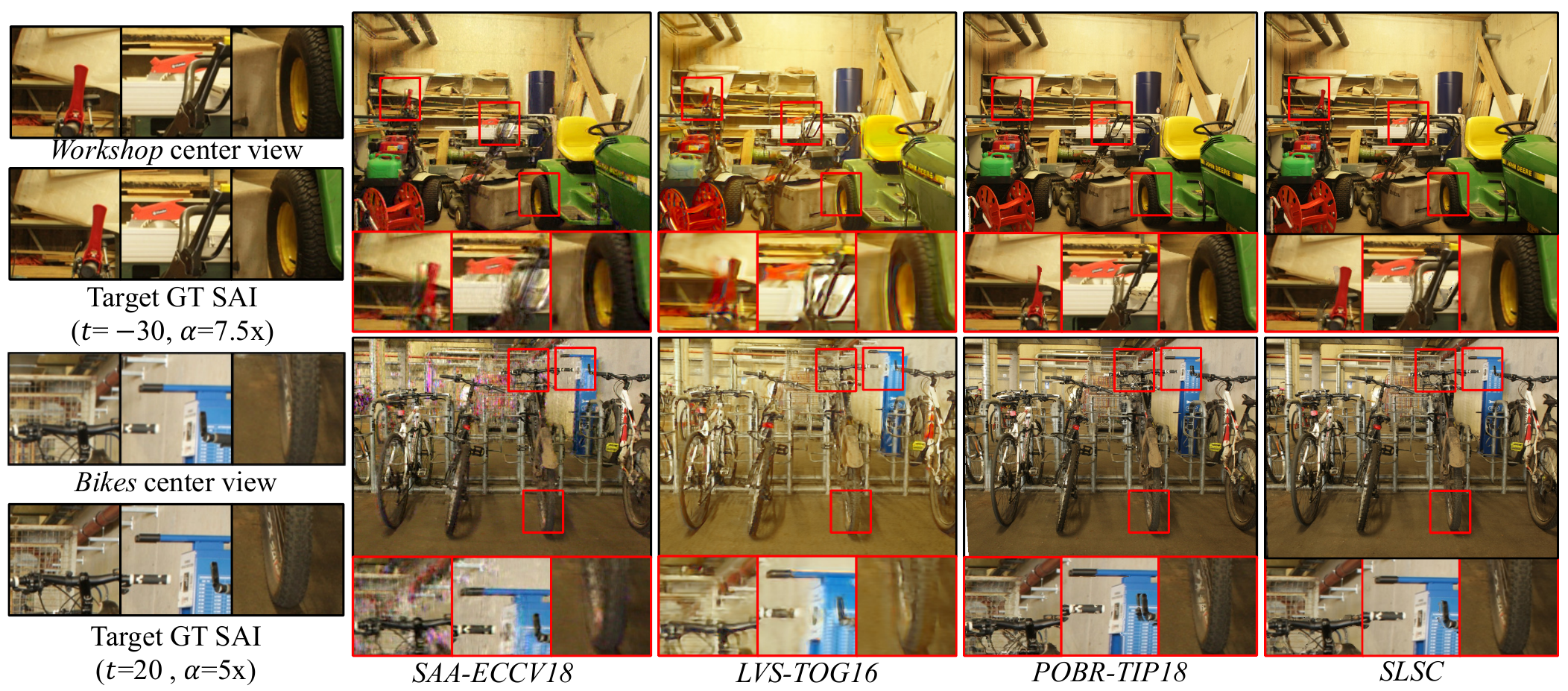}
	\end{center}
	\vspace{-0.6cm}
	\caption{Comparison of visual quality for synthesized target SAIs for the testing data \textit{Workshop} ($t$=-30, $\alpha$=7.5x), and \textit{Bikes} ($t$=20, $\alpha$=5x) \cite{Vamsi2017} from different methods.}
	\label{fig_visualComp1}
\end{figure*}
%%%%%%%%%%%%%%%%%%%%%%%Figure%%%%%%%%%%%%%%%%%%%%%%%

\subsection{Qualitative Evaluation}

We visually compare the quality of rendered novel views from different methods. 
Fig. \ref{fig_pcocOutput} shows the output from the \textit{PCOC}-Net based on the input features. We can see that the ambiguous regions caused by occlusions and perspective shifting have been restored well. In the top-left example, note how the width of the chair legs have been restored to normal compared with the distorted input $V_d^t$. Note that to directly assess the \textit{PCOC} network, we have not added any post-processing algorithms over the inpainted areas. The \textit{PCOC} network learns the textures of missing areas, and other simple techniques can be applied to blend the color to be consistent with its surroundings \cite{iizuka2017globally}. 

More visual comparisons are given in Fig. \ref{fig_visualComp1}. As can be seen that the rendered novel views from our proposed \textit{SLSC} model provides the highest quality outputs, while other competing methods show obvious blur and errors for structures with significant angular transforms.

\subsection{Ablation Study}
For the ablation study, we focus on the contribution of the surface consistent label map $\tilde{W}_L^t$ on the quality of synthesized novel SAIs. 
With $\tilde{W}_L^t$ as guide, the \textit{SLSC} system produces more stable and surface consistent inpainting outcomes as highlighted in Fig. \ref{fig_scInfluence}. The baseline system \textit{SL} (without \textit{SC}) tends to produce blurred and incorrect predictions around the occlusion borders.

% \subsection{Discussion}

% LF baseline extension is a challenging problem. The key to high quality synthesis is an accurate estimation of scene depth. It is challenging for even state-of-the-art methods at intricate and/or textureless regions. LF view synthesis modeled re-trained for this task proves to perform poorly mainly because the content variation is much more drastic than view interpolation problems, and it requires accurate estimation of scene structures in order to preserve it well from another drastically different viewing angle. We believe our stratified analysis methodology has high potential for further improvement when it is combined with of structure level understanding of scene geometry.

\section{Conclusion}\label{sec_conclusion}
\vspace{-0.2cm}
We have proposed an LF view synthesis algorithm which renders high quality novel LF views far outside the angular baselines of reference SAIs. A Stratified synthesis strategy is adopted which parses the scene content based on different disparity layers and based on different spatial granularities. Such a stratified methodology proves to be helpful in preserving scene content structures over large perspective shifts, and it provides informative clues for inferring the textures of occluded regions. A generative-adversarial model has been adopted for parallax correction and occlusion completion. Experiments show that our proposed model can provide more reliable novel view synthesis quality especially at large baseline extension ratios.

{\small
	\bibliographystyle{ieee}
	\bibliography{mybib}

\begin{thebibliography}{10}\itemsep=-1pt

\bibitem{achanta2012slic}
R.~Achanta, A.~Shaji, K.~Smith, A.~Lucchi, P.~Fua, and S.~Susstrunk.
\newblock Slic superpixels compared to state-of-the-art superpixel methods.
\newblock {\em IEEE Transactions on Pattern Analysis and Machine Intelligence},
  34(11):2274--2282, 2012.

\bibitem{Vamsi2017}
V.~K. {Adhikarla}, M.~{Vinkler}, D.~{Sumin}, R.~K. {Mantiuk}, K.~{Myszkowski},
  H.~{Seidel}, and P.~{Didyk}.
\newblock Towards a quality metric for dense light fields.
\newblock In {\em IEEE Conference on Computer Vision and Pattern Recognition},
  pages 3720--3729, July 2017.

\bibitem{chen2018light}
J.~Chen, J.~Hou, and L.~P. Chau.
\newblock Light field compression with disparity-guided sparse coding based on
  structural key views.
\newblock {\em IEEE Transactions on Image Processing}, 27(1):314--324, Jan
  2018.

\bibitem{chen2018accurate}
J.~{Chen}, J.~{Hou}, Y.~{Ni}, and L.~{Chau}.
\newblock Accurate light field depth estimation with superpixel regularization
  over partially occluded regions.
\newblock {\em IEEE Transactions on Image Processing}, 27(10):4889--4900, Oct
  2018.

\bibitem{glorot2010understanding}
X.~Glorot and Y.~Bengio.
\newblock Understanding the difficulty of training deep feedforward neural
  networks.
\newblock In {\em International Conference on Artificial Intelligence and
  Statistics}, pages 249--256, 2010.

\bibitem{goodfellow2014generative}
I.~Goodfellow, J.~Pouget-Abadie, M.~Mirza, B.~Xu, D.~Warde-Farley, S.~Ozair,
  A.~Courville, and Y.~Bengio.
\newblock Generative adversarial nets.
\newblock In {\em Advances in neural information processing systems}, pages
  2672--2680, 2014.

\bibitem{he1998layered}
L.-w. He, J.~Shade, S.~Gortler, and R.~Szeliski.
\newblock Layered depth images.
\newblock 1998.

\bibitem{hou2018light}
J.~Hou, J.~Chen, and L.-P. Chau.
\newblock Light field image compression based on bi-level view compensation
  with rate-distortion optimization.
\newblock {\em IEEE Transactions on Circuits and Systems for Video Technology},
  29(2):517--530, 2019.

\bibitem{Huang2015LFStereo}
F.-C. Huang, K.~Chen, and G.~Wetzstein.
\newblock The light field stereoscope: Immersive computer graphics via factored
  near-eye light field displays with focus cues.
\newblock {\em ACM Transactions on Graphics}, 34(4):60:1--60:12, July 2015.

\bibitem{huang2015light}
F.-C. Huang, K.~Chen, and G.~Wetzstein.
\newblock The light field stereoscope: immersive computer graphics via factored
  near-eye light field displays with focus cues.
\newblock {\em ACM Transactions on Graphics (TOG)}, 34(4):60, 2015.

\bibitem{iizuka2017globally}
S.~Iizuka, E.~Simo-Serra, and H.~Ishikawa.
\newblock Globally and locally consistent image completion.
\newblock {\em ACM Transactions on Graphics}, 36(4):107, 2017.

\bibitem{isola2017image}
P.~Isola, J.-Y. Zhu, T.~Zhou, and A.~A. Efros.
\newblock Image-to-image translation with conditional adversarial networks.
\newblock {\em arXiv preprint}, 2017.

\bibitem{jeon2015accurate}
H.-G. Jeon, J.~Park, G.~Choe, J.~Park, Y.~Bok, Y.-W. Tai, and I.~So~Kweon.
\newblock Accurate depth map estimation from a lenslet light field camera.
\newblock In {\em IEEE Conference on Computer Vision and Pattern Recognition},
  pages 1547--1555, 2015.

\bibitem{kalantari2016learning}
N.~K. Kalantari, T.-C. Wang, and R.~Ramamoorthi.
\newblock Learning-based view synthesis for light field cameras.
\newblock {\em ACM Transactions on Graphics}, 35(6):193, 2016.

\bibitem{Keys1981cubic}
R.~{Keys}.
\newblock Cubic convolution interpolation for digital image processing.
\newblock {\em IEEE Transactions on Acoustics, Speech, and Signal Processing},
  29(6):1153--1160, December 1981.

\bibitem{Kim2013}
C.~Kim, H.~Zimmer, Y.~Pritch, A.~Sorkine-Hornung, and M.~H. Gross.
\newblock Scene reconstruction from high spatio-angular resolution light
  fields.
\newblock {\em ACM Transactions on Graphics}, 32(4):73--1, 2013.

\bibitem{kingma2014adam}
D.~Kingma and J.~Ba.
\newblock Adam: A method for stochastic optimization.
\newblock {\em arXiv preprint arXiv:1412.6980}, 2014.

\bibitem{li2015superpixel}
Z.~Li and J.~Chen.
\newblock Superpixel segmentation using linear spectral clustering.
\newblock In {\em IEEE Conference on Computer Vision and Pattern Recognition},
  pages 1356--1363, June 2015.

\bibitem{lippmann1908la}
G.~Lippmann.
\newblock La photographie int\'{e}grale.
\newblock {\em Academie des Sciences}, 146:446--451, 1908.

\bibitem{Liu2005motion}
C.~Liu, A.~Torralba, W.~T. Freeman, F.~Durand, and E.~H. Adelson.
\newblock Motion magnification.
\newblock {\em ACM Transactions on Graphics}, 24(3):519--526, July 2005.

\bibitem{Ng2005}
R.~Ng, M.~Levoy, M.~Br{\'e}dif, G.~Duval, M.~Horowitz, and P.~Hanrahan.
\newblock Light field photography with a hand-held plenoptic camera.
\newblock {\em Computer Science Technical Report CSTR}, 2005.

\bibitem{ni2018reflection}
Y.~{Ni}, J.~{Chen}, and L.~{Chau}.
\newblock Reflection removal on single light field capture using focus
  manipulation.
\newblock {\em IEEE Transactions on Computational Imaging}, 4(4):562--572, Dec
  2018.

\bibitem{perra2016analysis}
C.~Perra, F.~Murgia, and D.~Giusto.
\newblock An analysis of {3D} point cloud reconstruction from light field
  images.
\newblock In {\em IEEE International Conference on Image Processing Theory
  Tools and Application}, pages 1--6, 2016.

\bibitem{perwass2012single}
C.~Perwass and L.~Wietzke.
\newblock Single lens 3d-camera with extended depth-of-field.
\newblock In {\em Human Vision and Electronic Imaging}, volume~17, pages
  829108--829108--15, 2012.

\bibitem{Shade1998layered}
J.~Shade, S.~Gortler, L.-w. He, and R.~Szeliski.
\newblock Layered depth images.
\newblock In {\em Annual Conference on Computer Graphics and Interactive
  Techniques}, SIGGRAPH '98, pages 231--242, New York, NY, USA, 1998. ACM.

\bibitem{shin2018epinet}
C.~Shin, H.-G. Jeon, Y.~Yoon, I.~So~Kweon, and S.~Joo~Kim.
\newblock Epinet: A fully-convolutional neural network using epipolar geometry
  for depth from light field images.
\newblock In {\em IEEE Conference on Computer Vision and Pattern Recognition},
  pages 4748--4757, 2018.

\bibitem{stern2014perceivable}
A.~Stern, Y.~Yitzhaky, and B.~Javidi.
\newblock Perceivable light fields: Matching the requirements between the human
  visual system and autostereoscopic 3-d displays.
\newblock {\em Proceedings of the IEEE}, 102(10):1571--1587, Oct 2014.

\bibitem{tao2016depth}
M.~W. Tao, J.-C. Su, T.-C. Wang, J.~Malik, and R.~Ramamoorthi.
\newblock Depth estimation and specular removal for glossy surfaces using point
  and line consistency with light-field cameras.
\newblock {\em IEEE Transactions on Pattern Analysis and Machine Intelligence},
  38(6):1155--1169, 2016.

\bibitem{bergh2012seeds}
M.~Van~den Bergh, X.~Boix, G.~Roig, B.~de~Capitani, and L.~Van~Gool.
\newblock {\em {SEEDS:} Superpixels Extracted via Energy-Driven Sampling},
  pages 13--26.
\newblock Springer Berlin Heidelberg, 2012.

\bibitem{wang2015occlusion}
T.-C. Wang, A.~A. Efros, and R.~Ramamoorthi.
\newblock Occlusion-aware depth estimation using light-field cameras.
\newblock In {\em IEEE International Conference on Computer Vision}, pages
  3487--3495, 2015.

\bibitem{wang2016occlusion}
T.-C. Wang, A.~A. Efros, and R.~Ramamoorthi.
\newblock Depth estimation with occlusion modeling using light-field cameras.
\newblock {\em IEEE Transactions on Pattern Analysis and Machine Intelligence},
  38(11):2170--2181, 2016.

\bibitem{wang2017high}
T.-C. Wang, M.-Y. Liu, J.-Y. Zhu, A.~Tao, J.~Kautz, and B.~Catanzaro.
\newblock High-resolution image synthesis and semantic manipulation with
  conditional gans.
\newblock {\em arXiv preprint arXiv:1711.11585}, 2017.

\bibitem{wang20164d}
T.-C. Wang, J.-Y. Zhu, E.~Hiroaki, M.~Chandraker, A.~A. Efros, and
  R.~Ramamoorthi.
\newblock A 4d light-field dataset and cnn architectures for material
  recognition.
\newblock In {\em European Conference on Computer Vision}, pages 121--138.
  Springer, 2016.

\bibitem{wang2017light}
T.-C. Wang, J.-Y. Zhu, N.~K. Kalantari, A.~A. Efros, and R.~Ramamoorthi.
\newblock Light field video capture using a learning-based hybrid imaging
  system.
\newblock {\em ACM Transactions on Graphics}, 36(4):133, 2017.

\bibitem{wanner2014variational}
S.~Wanner and B.~Goldluecke.
\newblock Variational light field analysis for disparity estimation and
  super-resolution.
\newblock {\em IEEE transactions on pattern analysis and machine intelligence},
  36(3):606--619, 2014.

\bibitem{wu2017light}
G.~Wu, M.~Zhao, L.~Wang, Q.~Dai, T.~Chai, and Y.~Liu.
\newblock Light field reconstruction using deep convolutional network on epi.
\newblock In {\em Proceedings of the IEEE Conference on Computer Vision and
  Pattern Recognition}, pages 6319--6327, 2017.

\bibitem{yeh2017semantic}
R.~A. Yeh, C.~Chen, T.-Y. Lim, A.~G. Schwing, M.~Hasegawa-Johnson, and M.~N.
  Do.
\newblock Semantic image inpainting with deep generative models.
\newblock In {\em CVPR}, volume~2, page~4, 2017.

\bibitem{yeung2018fast}
H.~W.~F. Yeung, J.~Hou, J.~Chen, Y.~Ying~Chung, and X.~Chen.
\newblock Fast light field reconstruction with deep coarse-to-fine modeling of
  spatial-angular clues.
\newblock In {\em European Conference on Computer Vision}, pages 137--152,
  2018.

\bibitem{Zhou2018stereo}
T.~Zhou, R.~Tucker, J.~Flynn, G.~Fyffe, and N.~Snavely.
\newblock Stereo magnification: Learning view synthesis using multiplane
  images.
\newblock {\em ACM Transactions on Graphics}, 37(4):65:1--65:12, July 2018.

\end{thebibliography}
}

\end{document}